\documentclass{article} 
\usepackage{colm2024_conference}

\usepackage{microtype}

\usepackage{tabularx}
\usepackage{makecell}
\usepackage{algpseudocode}
\usepackage{booktabs}
\usepackage[export]{adjustbox}
\usepackage{hhline}
\usepackage{xspace}
\usepackage{xcolor}         
\usepackage{url}
\usepackage{graphicx}
\usepackage{wrapfig}
\usepackage{trajan}
\usepackage{capt-of}
\usepackage{multicol}
\usepackage{multirow}
\usepackage{inconsolata}
\usepackage{listings}
\usepackage{todonotes}
\usepackage[cachedir=.,frozencache]{minted}

\usepackage{graphicx}

\usepackage{verbatim}
\usepackage[labelformat=simple]{subcaption}

\usepackage{hyperref}

\definecolor{Box1Color}{RGB}{227, 236, 246}
\definecolor{Box2Color}{RGB}{248, 220, 225}
\definecolor{Box3Color}{RGB}{255, 238, 224}

\newcommand{\modelpt}{\textsc{LLaVA-Hound-PT}\xspace}
\newcommand{\modelname}{\textsc{LLaVA-Hound-DPO}\xspace}
\newcommand{\modelsft}{\textsc{LLaVA-Hound-SFT}\xspace}

\newcommand{\datasetname}{\textsc{ShareGPTVideo}\xspace}

\newcommand{\webvid}{WebVid\xspace}
\newcommand{\vidal}{VIDAL\xspace}
\newcommand{\actnet}{ActivityNet\xspace}
\newcommand{\msrvtt}{MSRVTT\xspace}
\newcommand{\msvd}{MSVD\xspace}
\newcommand{\tgif}{TGIF\xspace}
\newcommand{\ssvt}{SSV2\xspace}

\newcommand{\pnum}{900k\xspace}

\newcommand{\videollava}{{Video-LLaVA}\xspace}
\newcommand{\gptv}{{GPT-4V}\xspace}
\newcommand{\chatgpt}{{ChatGPT}\xspace}
\newcommand{\gptf}{{GPT-4}\xspace}

\usepackage{amsmath,amsfonts,bm}

\def\eqref#1{equation~\ref{#1}}

\def\1{\bm{1}}

\DeclareMathAlphabet{\mathsfit}{\encodingdefault}{\sfdefault}{m}{sl}
\SetMathAlphabet{\mathsfit}{bold}{\encodingdefault}{\sfdefault}{bx}{n}

\newcommand{\sigmoid}{\sigma}

\usepackage{cleveref}

\title{Direct Preference Optimization of Video Large Multimodal Models from Language Model Reward}

\newcommand{\cmu}{\spadesuit}
\newcommand{\bytedance}{\diamondsuit}
\newcommand{\austin}{\nabla}
\newcommand{\columbia}{\triangle}
\newcommand{\ntu}{\heartsuit}

\author{
\hspace{0.15\linewidth} Ruohong Zhang$^{*\cmu}$\hspace{5em} Liangke Gui\thanks{Equal contribution.} 
\ $^{\cmu\bytedance}$ \\[5pt]
\hspace{1em}\textbf{Zhiqing Sun$^{\cmu}$\,, Yihao Feng$^{\austin}$\,, Keyang Xu$^\columbia$\,, Yuanhan Zhang$^{\ntu}$\,, Di Fu$^{\bytedance}$}\,,
 \\ 
 \hspace{1em}\textbf{Chunyuan Li$^{\bytedance}$\,, Alexander Hauptmann$^{\cmu}$\,, Yonatan Bisk$^{\cmu}$\,,  Yiming Yang$^{\cmu}$}
 \\
\hspace{1em} $^\cmu$CMU LTI, $^\bytedance$Bytedance,
 $^\austin$UT Austin,
 $^{\columbia}$Columbia University, $^{\ntu}$NTU \\[5pt]
 \hspace{0.08\linewidth}\texttt{Project Page: \url{https://github.com/RifleZhang/LLaVA-Hound-DPO}}
}

\colmfinalcopy 
\begin{document}

\maketitle

\begin{abstract}

Preference modeling techniques, such as direct preference optimization (DPO), has shown effective in enhancing the generalization abilities of large language model (LLM). However, in tasks involving video instruction-following, providing informative feedback, especially for detecting hallucinations in generated responses, remains a significant challenge. Previous studies have explored using large large multimodal models (LMMs) as reward models to guide preference modeling, but their ability to accurately assess the factuality of generated responses compared to corresponding videos has not been conclusively established. This paper introduces a novel framework that utilizes detailed video captions as a proxy of video content, enabling language models to incorporate this information as supporting evidence for scoring video Question Answering (QA) predictions. Our approach demonstrates robust alignment with OpenAI \gptv model's reward mechanism, which directly takes video frames as input. Furthermore, we show that applying this tailored reward through DPO significantly improves the performance of video LMMs on video QA tasks.

\end{abstract}

\section{Introduction}
\label{sec:intro}


This paper addresses the challenge of aligning LMMs, particularly in tasks that involve video instruction following. Despite recent advancements in reinforcement learning (RL)~\citep{ouyang2022training, bai2022constitutional, lee2023rlaif, sun2023salmon} and DPO~\citep{rafailov2024direct, chen2024self, hosseini2024v}, which have been effective in guiding LLMs towards generating more honest, helpful, and harmless content, their effectiveness in multimodal contexts remains limited. The critical obstacle lies in developing a robust reward system capable of distinguishing preferred responses from less preferred ones, especially when such responses are generated based on video inputs. The challenge is further complicated by the presence of hallucinations in generated content, stemming from the scarcity of alignment data across different modalities~\citep{liu2023visual, sun2023aligning}.

While human preference data is valuable, it is challenging to scale due to its cost and labor-intensive nature, as highlighted by the LLaVA-RLHF~\citep{sun2023aligning} paper, which collected 10k human-evaluated instances at a considerable cost of \$3000. Existing approaches for distlling preferences, such as those for image data using \gptv~\citep{li2023silkie}, encounter scalability issues, especially for video inputs that require analyzing multiple frames. While \cite{ahn2024tuning} leverage a supervised finetuning (SFT) model for self-evaluation, the efficacy of the SFT model remains uncertain, particularly in accurately assessing the factuality of responses in relation to their corresponding videos.

To tackle the aforementioned challenges, we introduce a cost-effective reward mechanism aimed at reliably evaluating the quality of responses generated by video (LLMs), serving as a basis for further preference optimization. 
We propose the use of detailed video captions as a proxy for video content, enabling a language model analyze video content and assess the accuracy of an LMM's response to a related question and determine the presence of hallucinations. The language model provides natural language feedback as a chain-of-thought step, and generates a numerical score for reward, facilitating a cost-effective feedback system. 

However, high-quality video captions are essential for this process. To mitigate the shortage of high-quality video captions, we have developed a comprehensive video caption dataset, \datasetname, using a novel prompting technique with the \gptv model, comprising 900k captions that encompass a wide range of video content, including temporal dynamics, world knowledge, object attributes, and spatial relationships.
With this video caption dataset available, we verify that our reward mechanism, which utilizes video captions as a proxy, is well-aligned with evaluations derived from the more powerful, albeit costlier, \gptv model-generated rewards. Employing this reward mechanism as the basis for DPO algorithm, we train \modelname that achieves an 8.1\% accuracy improvement over the SFT counterpart. This marks a significant advancement in video LMM alignment and represents the first successful application of a DPO method in this domain.

Our contributions are outlined as follows:

\begin{enumerate}
\item We develop a large-scale, detailed video caption dataset, covering a wide array of content. This dataset serves as a foundational resource for LMM model training and research, facilitating advancements in video understanding tasks.
    \item We introduce a cost-effective method for evaluating video instruction-following tasks, serving as enhanced evaluation of model performance. 
    \item We demonstrate the effective application of DPO to improve model performance by leveraging the language model feedback as reward, which substantially improves the alignment of video LMM, establishing a new benchmark for SOTA performance in video QA tasks.
\end{enumerate}

\section{Related Work}
\subsection{Large Multi-Modal Models}
LMMs ~\citep{liu2023visual, liu2023improved, bai2023qwen, chen2023shikra, li2023blip} have enabled instruction following across modalities by utilizing LLM as backbones. In the context of video understanding, LLMs have been adapted to process video content~\citep{lin2023videollava, zhang2023video, maaz2023video, li2023videochat, luo2023valley, liu2023one, jin2024video, ahn2024tuning}. Our work adots \videollava backbone, focusing on model enhancement through preference modeling with the DPO technique.

\subsection{Video-text Datasets}
Existing video-text datasets typically provide brief sentences or mere keywords as captions, as indicated by \cite{bain2021frozen, wang2023internvid, yu2019activitynet, jang2017tgif, xu2016msr}. \cite{shvetsova2023howtocaption} uses automatic speech recognition to extract textual content from videos, but it encounters alignment issues when the audio does not match or is absent from the visual content. Video-ChatGPT \citep{li2023videochat} employs human effort to create high-quality video instructions, albeit limited to the ActivityNet domain with only 100k instruction pairs. Our work leverages the \gptv model with specifically crafted prompts to produce detailed video captions as community resource for LMM training.

\subsection{Preference Modeling for LMMs}
Preference modeling techniques are employed to enhance the utility of LMMs while mitigating the issue of hallucination. \cite{sun2023aligning}  leveraged Reinforcement Learning with Human Feedback (RLHF) and incorporated caption information into the reward model to improve the assessment of factuality. More recently, \cite{ahn2024tuning} used RL on AI feedback to improve video LMM performance. For the image understanding, \cite{li2023silkie, gunjal2023detecting} introduced the application of DPO on the distilled rewards from \gptv on a group of model outputs, while \cite{zhao2023beyond} created preference data using \chatgpt to generate positive and negative pairs informed by detailed image descriptions. Our contribution extends DPO to the video LMM alignment, with the use of detailed captions as factual evidence for reward modeling.

\begin{figure}[t!]
    \centering
    \includegraphics[width=\linewidth]{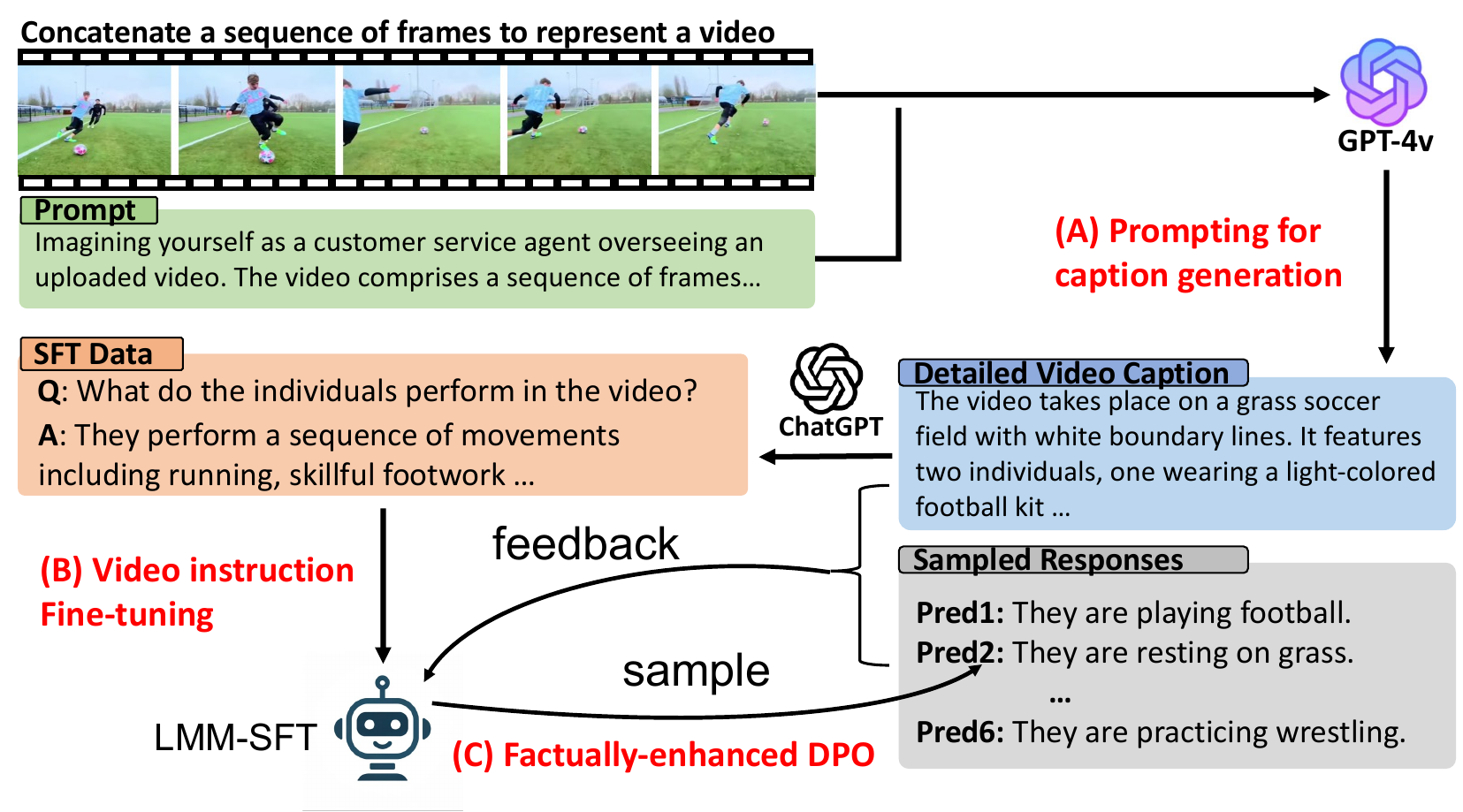}
    \caption{Workflow diagram showing: a) the use of \gptv for creating a detailed caption dataset for videos; b) generating video instruction data for SFT; c) integrating captions into a feedback loop for factually-enhanced DPO, improving the model's performance on video instruction-following tasks.}
    \label{fig:illustration}
\end{figure}

\section{Method}
\label{sec:method}

As shown in \cref{fig:illustration}, our methodology enhances video LMM alignment through DPO method using rewards from a language model. We elaborate on constructing a video caption dataset in \cref{subsec:caption_distillation}. Subsequently, in \cref{subsec:sft}, we discuss the generation of video instruction data and the fine-tuning process of our model. Lastly, \cref{subsec:dpo} details the incorporation of generated captions as a feedback mechanism for DPO method to refine our model's factual alignment in video instruction-following tasks.

\subsection{Prompting \gptv Model for Detailed Video Caption Distillation}
\label{subsec:caption_distillation}
The selection of dataset includes videos from three sources: the \webvid and \vidal datasets, which are general domain videos sourced from YouTube with 400k and 450k sampled videos respectively, and the \actnet dataset, which adds 50k videos focusing on human activities. The three datasets together result in a comprehensive collection of \pnum videos. To accommodate the requirement that \gptv only takes images as input, we preprocess videos by uniformly extracting ten frames per video content. These frames are then concatenated into a sequence to serve as a proxy for the video. This sequence is the input into \gptv to generate a coherent caption for the represented video based on the frame sequence. The prompt adheres to guidelines covering temporal dynamics, world knowledge, object attributes, spatial relationships, aesthetic assessments, etc., with the goal of comprehensively understanding the video contents.

\subsection{SFT with Generated Video Instruction Data from Detailed Caption}
\label{subsec:sft}
To generate video instruction-following data for SFT, we adopt a similar methodology outlined in Video-ChatGPT~\citep{li2023videochat}. Specifically, we first randomly sample 20k, 30k, 30k captions in our dataset from \actnet, \webvid and \vidal respective and then employ \chatgpt to generate three question-answer pairs given each detailed video caption, resulting in a total of 240k instruction data for finetuning. This approach ensures that the instructional data remains factually consistent with the content of the detailed captions. The specific prompting strategy used for this instruction generation process is detailed in \cref{fig:chatgpt_instruction_generation}.

\begin{figure}[t!]
    \centering
    \includegraphics[width=\linewidth]{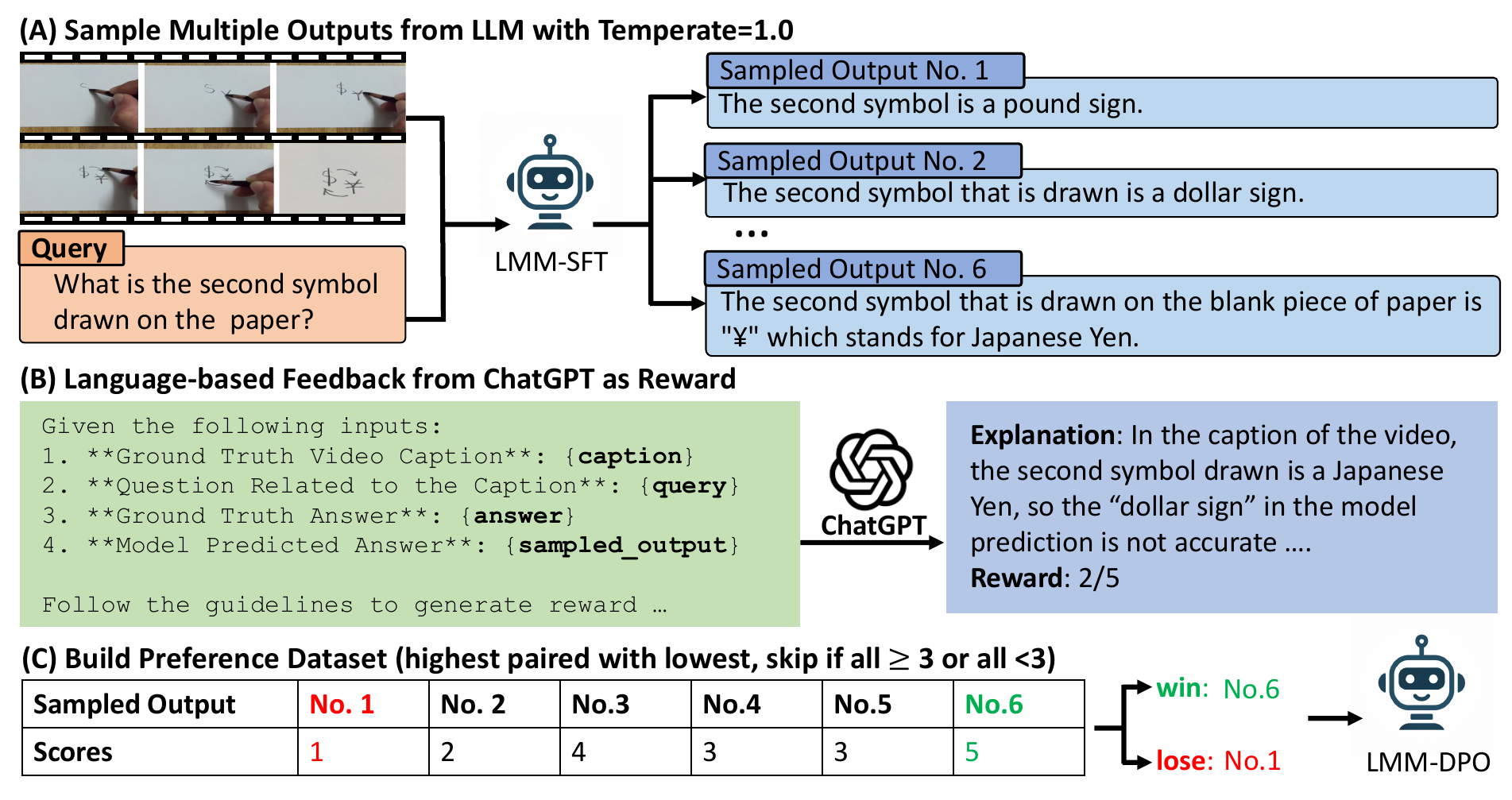}
    \caption{Detailed illustration of the proposed factually-enhanced DPO method.}
    \label{fig:dpo_illustration}
\end{figure}

\subsection{DPO with Feedback from Language Model as Reward}
\label{subsec:dpo}
Acquiring high-quality preference data is both costly and labor-intensive. Although \gptv is an effective model for reward distillation, its high cost, slow performance, and limited accessibility hinder scalability, especially for video inputs with multiple frames. We propose a cost-efficient method to generate reward data for DPO using detailed video captions as supporting evidence, as shown in \cref{fig:dpo_illustration}.

Initially, we randomly select a subset of 20k instruction pairs from the dataset described in \cref{subsec:sft}. The SFT model uses these sampled questions and their corresponding videos to generate six responses per input pair at a temperature of $1.0$. This procedure results in 120k question-answer pairs, which will be evaluated. Subsequently, we employ \chatgpt to process inputs including a question, the ground truth answer, the model's prediction, and a detailed description serving as supportive evidence, with the prompt in \cref{fig:chatgpt_verifier}. This generates an output that includes a natural language explanation as chain-of-thought step, followed by a numerical reward score on a scale from $1$ to $5$, indicating the level of factual alignment and overall quality.

For each video and question pair, we randomly select an answer with a score $\ge$ 3 as positive example, and an answer with a score below $3$ as negative. Cases where all responses are uniformly scored above or below $3$ are excluded from the dataset. After the selection process, approximately 17k training instances are compiled for DPO training. Formally, the dataset is denoted as $\mathcal{D}_{DPO} = \{ (\mathcal{V}, x, y_w, y_l )\}$, where $\mathcal{V}$ is the video, $x$ is the question, $y_w$ and $y_l$ are the positive and negative responses. The DPO objective is defined as below:
\begin{equation*}
\resizebox{.95\hsize}{!}{$\mathcal{L}_{\mathrm{DPO}}\left(\pi_\theta ; \pi_{\mathrm{ref}}\right)=-\mathbb{E}_{\left(\mathcal{V}, x, y_w, y_l\right) \sim \mathcal{D}_{DPO}}\left[\log \sigma\left(\beta \log \frac{\pi_\theta\left(y_w \mid x,\mathcal{V} \right)}{\pi_{\text {ref }}\left(y_w \mid x,\mathcal{V}\right)}-\beta \log \frac{\pi_\theta\left(y_l \mid x,\mathcal{V}\right)}{\pi_{\text {ref }}\left(y_l \mid x,\mathcal{V}\right)}\right)\right]\,,$}
\end{equation*}
where $\pi_\theta$ is the policy model to be optimized and $\pi_{\text {ref }}$ is the base reference model, both models are initialized with SFT weights. $\sigmoid$ is the logistic function and $\beta$ is set to $0.1$.

Our approach to reward assignment leverages detailed captions as a proxy for video frames, offering both cost-effectiveness and efficiency. This method incurs costs of less than \$20, under a pricing model of \$1.5 per million tokens. In comparison, previous methods of preference data collection, such as in \cite{sun2023aligning}, required an expenditure of \$3,000 to gather 10k human preference data points. Additionally, the method proposed by \cite{li2023silkie}, which employs \gptv for reward data labeling, incurs a significantly higher cost—\$30 per million tokens—and demonstrates considerably slower inference speeds.

\begin{figure}[ht]
\centering
\includegraphics[width=0.45\linewidth, valign=t]{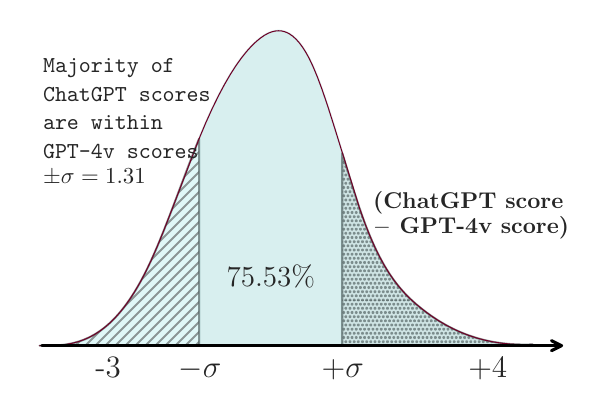} 
\includegraphics[width=0.45\linewidth, valign=t]{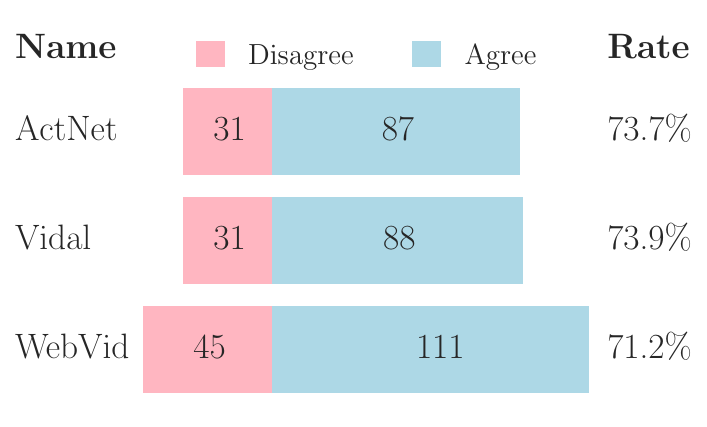} 
\caption{Assessing Evaluator Quality Using Captions in Place of Frames. The left figure shows the distribution of evaluation score differences between \chatgpt (with caption as proxy) and \gptv (directly on frames) evaluations. The right figure shows the rate of preference agreement between \chatgpt and \gptv as evaluators.}

\label{fig:evaluator_assessment}
\end{figure}

\section{Assessment of Evaluator with \gptv Caption as Evidence}
\label{sec:evaluator_assessment}


To assess the effectiveness of our proposed reward assignment method, which utilizes detailed captions as a proxy of actual video frames, we conducted a comparative analysis with the \gptv, used as a video QA evaluator. The latter reward system employs \gptv evaluation directly taking in video frames, a question, and the model prediction as inputs, with detailed prompt in \cref{fig:gptv_verifier}. Both reward systems follow the same set of guidelines for scoring reward. 

To compare the two methods, we sample $200$ videos from each of the \webvid, \vidal, and \actnet datasets, each associated with one question and two model predictions from our SFT model, with one preferred and one dispreferred by \chatgpt. This results in $1,200$ examples, for which we used \gptv (with the "gpt-4-vision-preview" version) version to assign scores. Filtering through the Azure API backend resulted in $196$, $151$, and $143$ videos from each dataset, respectively, having both answers evaluated. The average scores of all examples from \chatgpt and \gptv evaluations were $2.9$ and $3.5$ respectively, indicating a tendency of \gptv to yield slightly positive evaluations. The Pearson Correlation Coefficient (PCC) of $0.47$ ($p<0.01$) suggests a moderate positive correlation. In \cref{fig:evaluator_assessment} (left), the distribution of the difference between \chatgpt and \gptv scores reveals that majority ($>75\%$) of \chatgpt scores fall within one standard deviation ($\sigma=1.31$) of \gptv scores. Additionally, in \cref{fig:evaluator_assessment} (right), the agreement on preference between \chatgpt and \gptv, excluding ties, exceeded $70\%$. These findings cautiously support our benchmark's applicability in video QA evaluation. Further refinements for better alignment—such as incorporating Likert scales~\cite{zhou2023sotopia} or \gptf evaluation—are areas for future research.

\begin{table*}[!ht]
\centering
\resizebox{1.0\linewidth}{!}{
    \begin{tabular}{lccccccc}
    \toprule
    & & \multicolumn{6}{c}{\bf Existing Video QA Benchmark from \cite{maaz2023video}} \\
    \multirow{2}{*}{\textbf{Methods}}
    & \multirow{2}{*}{\textbf{LLM Size}}
    & \multicolumn{2}{c}{\textbf{MSVD-QA}}
    & \multicolumn{2}{c}{\textbf{MSRVTT-QA}}
    & \multicolumn{2}{c}{\textbf{TGIF-QA}}  \\
    \cmidrule(lr){3-4}                  
    \cmidrule(lr){5-6}
    \cmidrule(lr){7-8}
    &
    & Acc. & Score 
    & Acc. & Score 
    & Acc. & Score \\
    \midrule
    FrozenBiLM~\citep{yang2022zero}$*$      
            & 1B & 32.2 & - & 16.8  & - & 41.0 & - \\
    VideoLLaMA~\citep{zhang2023video}$*$
            & 7B & 51.6 & 2.5 & 29.6 & 1.8 & - & - \\
    LLaMA-Adapter~\citep{zhang2023llama}$*$
            & 7B & 54.9 & 3.1 & 43.8 & 2.7 & - & - \\
    VideoChat~\citep{li2023videochat}$*$
            & 7B
            & 56.3 & 2.8 & 45.0 & 2.5 & 34.4 & 2.3 \\
    BT-Adapter~\cite{liu2023one}$*$
            & 7B & 67.5 & 3.7 & 57.0 & 3.2 & - & - \\
    Video-ChatGPT~\citep{maaz2023video}
            & 7B & 68.6 & 3.8 & 58.9 & 3.4 & 47.8 & 3.2 \\
    Chat-UniVi~\citep{Chat-UniVi} & 7B & 70.0 & 3.8 & 53.1 & 3.1 & 46.1 & 3.1 \\
    VideoChat2~\cite{li2023mvbench}
            & 7B & 70.0 & 3.9 & 54.1 & 3.3 & - & - \\
    Video-LLaVA~\cite{lin2023video}
            & 7B & 71.8 & 3.9 & 59.0 & 3.4 & 48.4 & 3.2 \\
    LLaMA-VID~\citep{li2023llama}
            & 7B & 72.6 & 3.9 & 58.7 & 3.4 & 49.2 & 3.3 \\
    LLaMA-VID~\cite{li2023llama}
            & 13B & 74.3 & 4.0 & 59.8 & 3.4 & 50.8 & 3.3 \\
    VLM-RLAIF~\citep{ahn2024tuning}$*$ & 7B & 76.4 & 4.0 & 63.0 & 3.4 & - & - \\
    \midrule
    \modelsft  & 7B & 75.7 & 3.9 & 58.7 & 3.3 & 53.5 & 3.3 \\
    \modelname & 7B & \textbf{80.7} & \textbf{4.1} & \textbf{70.2} & \textbf{3.7} & \textbf{61.4} & \textbf{3.5} \\
    \bottomrule
    \hline
    \end{tabular}
}
\caption{
\textbf{Evaluation of Model Performance on Zero-Shot Video Question Answering Benchmarks Using gpt-3.5-turbo-0613.} Models denoted with $*$ have their results directly sourced from their original publications. Caution is advised when interpreting these results; see Appendix~\ref{apd:official_evaluation} for an in-depth analysis of evaluation challenges. All other baseline models were reproduced by our team. 
}
\label{tab:official_eval}
\end{table*}

\section{Experimental Results}
\subsection{Model Architecture, Image Data Mix-up and  Training Pipelines }
We adopt \videollava ~\citep{lin2023videollava} as the backbone of our video LMM, but our dataset and method can be applied to any other architectures as well. Specifically, \videollava employs LanguageBind ~\citep{zhu2023languagebind} encoder for image and video frame inputs, a MLP projector with 2 fully connected layers to map visual embeddings into text space, and Vicuna~\cite{chiang2023vicuna} as large language model. During training, we first initialize the projection MLP layer with the 
same \videollava MLP weight. Then we follow the training stages below:

\noindent \textbf{Caption Pre-training Stage (\modelpt):}
At pretraining stage, we use captioning data including 650k image caption data from ALLaVA ~\citep{chen2024allava} and our distilled 900k video caption. We freeze the LanguageBind visual encoder and fine-tune the MLP projector and LLM, with learning rate 2e-5 and batch size 128.

\noindent \textbf{SFT Stage (\modelsft):}
We use instructional data from both image and video domain to fine-tune the model for instruction-following ability. Our SFT model use
 600k image instruction data from ALLaVA and our generated 240k video instruction data, with learning rate 5e-6 and batch size 128.

\noindent \textbf{DPO training Stage (\modelname):}
We use the 17k preference data introduced in \cref{subsec:dpo} for DPO training. Following \cite{ivison2023camels}, we train our policy model for $3$ epochs with learning rate 5e-7, and a batch size of 128, resulting in roughly 420 training steps. All the experiments are performed on 8 A100 gpus.

\subsection{Existing Benchmark Evaluation}
\paragraph{Dataset and Testing Environment}
We evaluate model performance on three benchmark datasets: \msvd-QA~\cite{chen2011collecting}, \msrvtt-QA~\cite{xu2016msr}, and TGIF-QA~\cite{jang2017tgif}, using \chatgpt with version gpt-3.5-turbo-0611 to assess model predictions. The evaluation prompts follow  \cite{maaz2023video}. In our experiment, we found that different ChatGPT versions have high impact on absolute score of metric, but the overall ranking of models is relatively stable. We select gpt-3.5-turbo-0613 due to its closeness to the reported score in \videollava paper. Further details on the selection rationale and evaluation pitfalls are discussed in Appendix~\ref{apd:official_evaluation}.

\paragraph{Baseline Selection}
Our selection criteria include video LMM models that have demonstrated SOTA performance, specifically including \videollava, which is also our choice of architecture. We consider other contemporaneous SOTA models with similar reported performance levels to \videollava, yet have not been directly compared in prior studies. A key consideration in our selection is the availability of models with accessible code and checkpoints, which is crucial for ensuring reproducibility of our findings. To this end, we replicate models including Video-ChatGPT~\citep{maaz2023video}, LLaMA-VID~\citep{li2023llama} (7B and 13B), Chat-UniVi~\citep{Chat-UniVi}, and Video-LLaVA~\cite{lin2023video}. We adopt the results from additional baselines including FrozenBiLM~\citep{yang2022zero}, VideoChat~\citep{li2023videochat} and VideoLLaMA~\citep{zhang2023video}, sourced from their original publication. 

\begin{figure}[t!]
    \centering
    \vspace{-1cm}
    \includegraphics[width=0.9\textwidth]{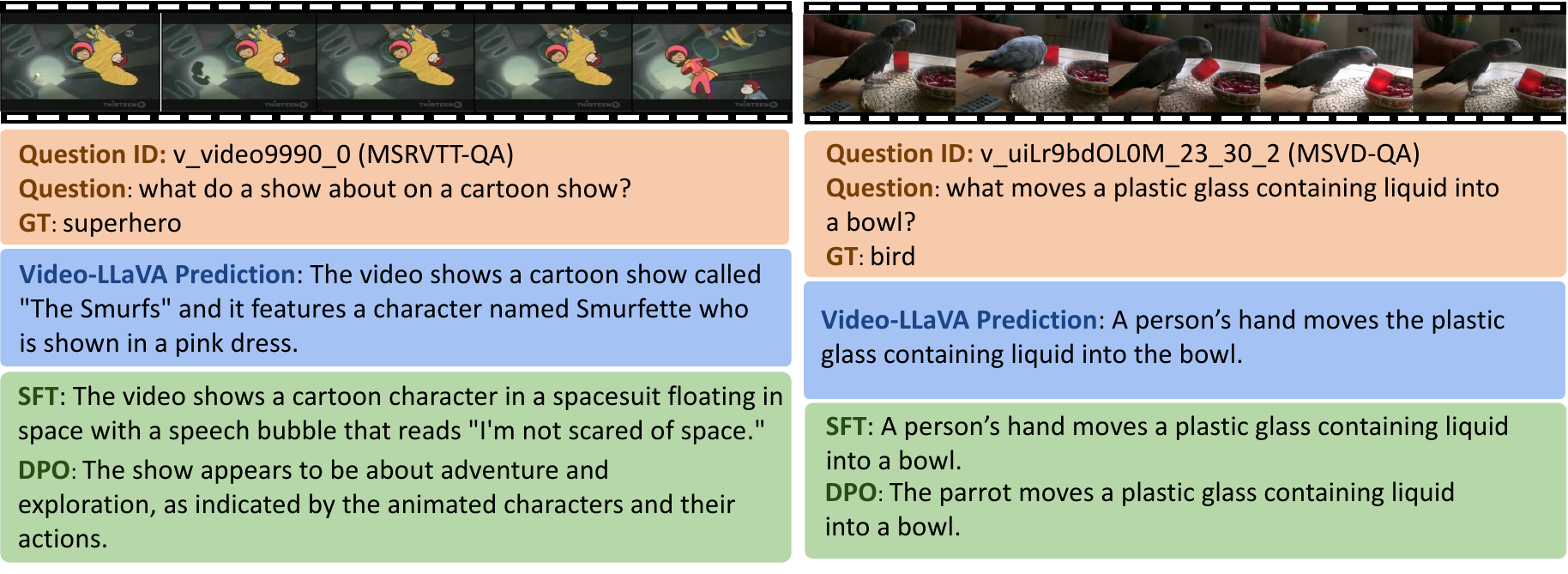}
    \caption{Examples from MSRVTT-QA and MSVD-QA showcase that our \modelname generates better responses, and reveal key limitations of the existing benchmark evaluation.}
    \label{fig:error-analysis}
\end{figure}

\paragraph{Results} In \cref{tab:official_eval}, our analysis shows that within the SFT models, LLaMA-VID-7B and \videollava exhibit comparable performance, with LLaMA-VID-13B performing the best. Our \modelsft model achieves comparable performance to LLaMA-VID-13B. Incorporating preference modeling, \modelname achieves an average accuracy of $70.75\%$, surpassing \modelsft, which has an average accuracy of $62.65\%$, by $8.1\%$. Furthermore, \modelname, enhanced by DPO, exhibits superior accuracy compared to VLM-RLAIF's performance achieved through reinforcement learning.

\paragraph{Error Analysis} \Cref{fig:error-analysis} illustrates two examples. In the left example, \modelsft provides an accurate description of the video's first half but introduces a hallucination with the phrase ``I'm not scared of space," absent in the video content. \modelname yields a more accurate inference. In the right example, both \modelsft and \videollava models produce incorrect inferences, whereas \modelname successfully correctly identifies the subject in the video. More critically, these examples unveil two significant issues within the current benchmark: (1) the auto-generated questions from existing benchmark may be grammatically incorrect or even nonsensical, and (2) the answers are limited to a \textit{single} word, which is insufficient for evaluating LMMs with long-form text generation. Such constraints in the ground truth answers hinder the evaluation of crucial aspects like helpfulness and hallucination detection.



\subsection{Proposed Benchmark Evaluation with \gptv Caption as Supporting Evidence}
As a solution to the above limitations in existing benchmark evaluation, we propose a new set of test questions for same videos in the benchmark datasets with generated QA from detailed captions, illustrated in \cref{apd:video_qa}. Applying the our reward system in \cref{sec:evaluator_assessment}, we report the score from \chatgpt, and a score value $\ge 3$ will be considered correct for accuracy calculation. This new long-form QA evaluation potentially support diverse aspects in responses including relevance, accuracy, clarity and completeness in prompt \ref{fig:gptv_verifier}.

\begin{table*}[!t]
\centering
\resizebox{1.0\linewidth}{!}{
    \begin{tabular}{clcccccc}
    \toprule
    & & \multicolumn{6}{c}{\bf Proposed Video QA Benchmark (In-domain)} \\
    \multirow{2}{*}{\textbf{No.}} & \multirow{2}{*}{\textbf{Methods}}
    & \multicolumn{2}{c}{\textbf{\actnet-QA}}
    & \multicolumn{2}{c}{\textbf{\vidal-QA}}
    & \multicolumn{2}{c}{\textbf{\webvid-QA}}  \\  
    \cmidrule(lr){3-4}                  
    \cmidrule(lr){5-6}
    \cmidrule(lr){7-8}
    & & Acc. & Score 
    & Acc. & Score 
    & Acc. & Score \\
    \midrule
    \text{[1]} & Video-ChatGPT~\citep{maaz2023video} &  34.17 & 2.19 & 29.35 & 2.10 & 38.88 & 2.27 \\
    \text{[2]} & LLaMA-VID-7B~\citep{li2023llama}  & 36.54 & 2.27 & 30.58 & 2.15 & 36.99 & 2.24 \\
    \text{[3]} & LLaMA-VID-13B~\citep{li2023llama}  & 37.33 & 2.29 & 32.50 & 2.18 & 39.73 & 2.30 \\
    \text{[4]} & Chat-UniVi~\citep{Chat-UniVi}  & 39.35 & 2.32 & 31.40 & 2.16 & 40.05 & 2.31 \\
    \text{[5]} & Video-LLaVA~\cite{lin2023video} & 41.35 & 2.38 & 34.30 & 2.24 & 42.47 & 2.39 \\
    \midrule
    \text{[6]} & \modelsft & 66.62 & 3.05 & 60.50 & 2.88 & 71.07 & 3.17 \\
    \text{[7]} & \modelname & \bf 76.62 & \bf 3.18 & \bf 70.06 & \bf 3.04 & \bf 79.82 & \bf 3.29  \\
    \hline \hline
    \text{[8]} & \modelpt + Image Inst. & 69.31 & 3.09 & 60.57 & 2.85 & 68.03 & 3.02 \\
    \text{[9]} & \modelpt + VChat & 67.34 & 3.02 & 62.33 & 2.89 & 68.98 & 3.00  \\
    \text{[10]} & \modelname + training MLP & 71.89 & 3.10 & 65.57 & 2.95 & 75.37 & 3.21 \\
    \text{[11]} & \modelsft + Self-play & 64.11 & 2.85 & 56.28 & 2.68 & 67.89 &2.95 \\ 
    \text{[12]} & \modelname w/ lr3e-7 & 71.13 & 3.08 & 64.90 & 2.92 & 73.25 & 3.17\\ 
    \bottomrule
    \hline
    \end{tabular}
}
\caption{
Our proposed video QA benchmark evaluation on in-domain dataset using gpt-3.5-turbo-0301, with detailed captions as supporting evidence.
}
\label{tab:indomain_eval}
\end{table*}

\begin{table*}[!t]
\centering
\resizebox{\linewidth}{!}{
    \begin{tabular}{lcccccccc}
    \toprule
    & \multicolumn{8}{c}{\bf Proposed Video QA Benchmark (Out-of-domain)} \\
    \multirow{2}{*}{\textbf{Methods}}
    & \multicolumn{2}{c}{\textbf{\msvd-QA}}
    & \multicolumn{2}{c}{\textbf{\msrvtt-QA}}
    & \multicolumn{2}{c}{\textbf{\tgif-QA}}  
    & \multicolumn{2}{c}{\textbf{\ssvt-QA}} \\  
    \cmidrule(lr){2-3}                  
    \cmidrule(lr){4-5}
    \cmidrule(lr){6-7}
    \cmidrule(lr){8-9}
    & Acc. & Score 
    & Acc. & Score 
    & Acc. & Score 
    & Acc. & Score \\
    \midrule
    Video-ChatGPT~\citep{maaz2023video} &  34.06 & 2.20 & 25.65 & 1.98 & 31.35 & 2.09 & 19.36 & 1.75 \\
    LLaMA-VID-7B~\citep{li2023llama}  & 34.14 & 2.21 & 25.02 & 1.99 & 27.18 & 2.00 & 22.16 & 1.84 \\
    LLaMA-VID-13B~\citep{li2023llama} & 35.81 & 2.25 & 26.34 & 2.02 & 27.58 & 2.01 & 21.98 & 1.83 \\
    Chat-UniVi~\citep{Chat-UniVi} & 35.61 & 2.23 & 25.89 & 2.01 & 33.23 & 2.13 & 20.59 & 1.79 \\
    Video-LLaVA~\cite{lin2023video} & 39.46 & 2.37 & 30.78 & 2.15 & 32.95 & 2.18 & 24.31 & 1.90 \\
    \midrule
    \modelsft & 66.99 & 3.09 & 57.82 & 2.85 & 66.13 & 3.07 & 35.07 & 2.23 \\
    \modelname & \bf 73.64 & \bf 3.12& \bf68.29 & \bf 2.98 & \bf 74.00 & \bf 3.12 & \bf 48.89 & \bf 2.53  \\
    \hline \hline
    \modelpt + Image Inst. & 65.19 & 2.96 & 48.66 & 2.52 & 53.83 & 2.62 & 29.60 & 2.04 \\
    \bottomrule
    \hline
    \end{tabular}
}
\caption{
Our proposed video QA benchmark evaluation on out-of-domain dataset using gpt-3.5-turbo-0301, with detailed captions as supporting evidence.
}
\label{tab:outdomain_eval}
\end{table*}

\Cref{tab:indomain_eval} and \cref{tab:outdomain_eval} shows the in-domain and out-of-domain evaluation. We use "gpt-3.5-turbo-0301" for evaluation as it is the same version for constructing DPO dataset. 
The model performance is more distinguishable from our evaluation with \videollava performing the best among the other baseline models. 

\textbf{Video LMM without Video Instruction:} [8] in \cref{tab:indomain_eval} is baseline trained with only image instruction fine-tuned on \modelpt, which achieves an average accuracy of $65.97\%$, comparable to the \modelsft model's $66.06\%$ in in-domain QA scenarios. However, its performance significantly drops in out-of-domain QA contexts ($49.32\%$ vs. $56.50\%$), suggesting that Video QA training could potentially enhance generalization capabilities. 

\textbf{Quality of Generated SFT:} [9] substitutes our generated video QA  with the Video-ChatGPT dataset for \videollava fine-tuning. A comparison between the findings of [9] and [6] reveals a marginal performance disparity of $0.2\%$ in average accuracy, indicating that the quality of our generated QA closely parallels that of the existing video QA datasets. Given the similar quality in SFT data, the large gain of [6] over [5] can be reasonably concluded from large-scale pre-training on video captions.

\textbf{Unfreeze MLP:} The comparison between [10] and [7] reveals a significant decrease in performance when the MLP is unfrozen during DPO training. Despite this drop, however, the performance remains superior to that of the SFT baseline.

\textbf{Smaller Learning Rate:} The comparison between [12] and [7] reveals that using a smaller learning rate of 3e-7 (vs. 5e-7) results in a decreasing of model performance. This highlights the future improvements by finding better hyperparameters.

\textbf{Self-Play vs. DPO:} \cite{chen2024self} introduced a self-play methodology for DPO training, which designates ground truth answers as preferred and model-generated responses as dispreferred. When comparing the results of [11] with those in [6], a notable decrease in accuracy by $3\%$ from the SFT model is observed, suggesting that self-play may be less effective for video LMM alignment, and introducing reward model is helpful.

\begin{figure}[ht]
\centering
\vspace{-0.6cm}
\includegraphics[width=0.49\linewidth]{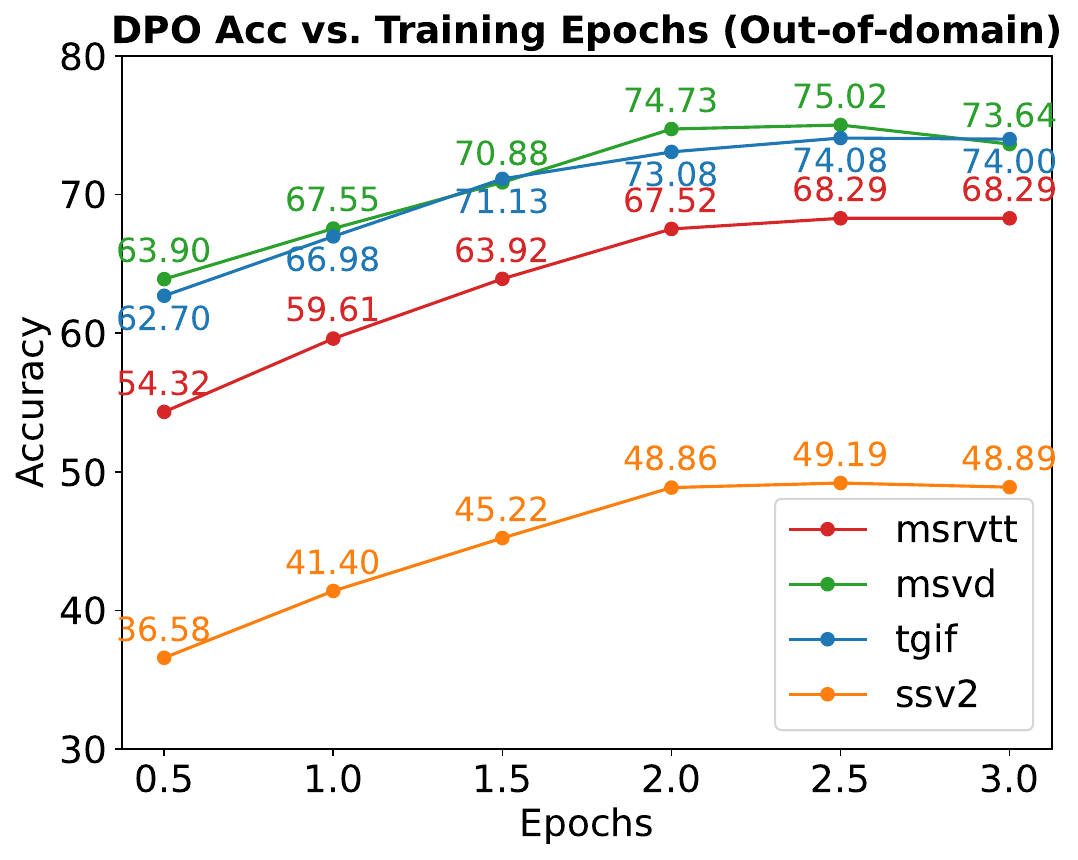} 
\includegraphics[width=0.49\linewidth]{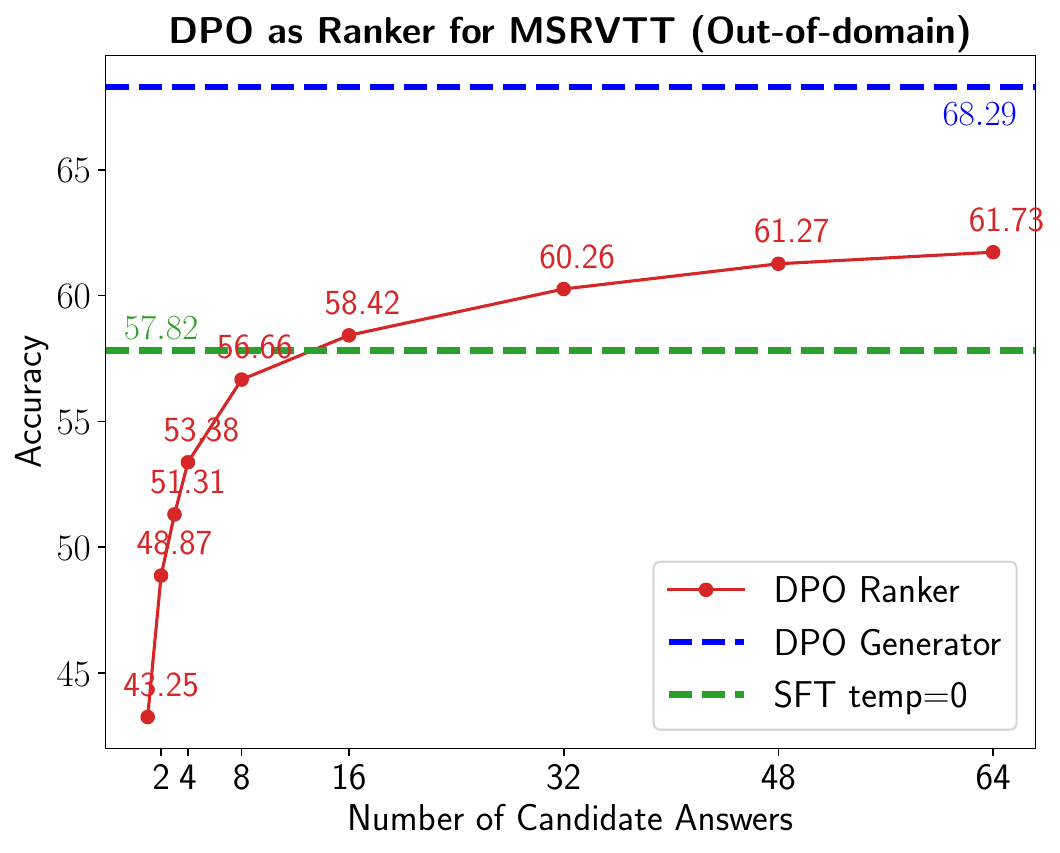} 
\caption{The left figure shows the test set accuracy of the DPO model w.r.t the number of training epochs. The right figure shows a comparison of DPO model performance as generator vs. ranker.}
\label{fig:dpo_ablation}
\end{figure}

\textbf{DPO Accuracy vs. Training Epochs.} The left of \cref{fig:dpo_ablation} depicts the generalization performance of the model on out-of-domain video QA tasks with respect to the number of training epochs. We observe a consistent enhancement in model performance among datasets during the initial 0 to 2 epochs, with peak performance materializing at around 2.5 epochs, which corresponds to 350 training steps.

\textbf{DPO as Ranker vs. Generator.}
Following \cite{hosseini2024v}, we compare the performance of employing the DPO model as a ranker for candidate answers produced by the SFT model, operating at a temperature setting of 1.0. As depicted on the right in \cref{fig:dpo_ablation}, we illustrate the test accuracy progression through the selection of the best among $N$ candidates by the DPO ranker. Initial observations indicate that the SFT model, when set to a temperature of 1.0, demonstrates a reduced accuracy (43.3\%) compared to that achieved through greedy decoding (57.8\%). A steady enhancement in performance is noted as the number of candidates increases, plateauing at an accuracy of approximately 62\% with 64 candidates. This performance, however, falls short when compared with the direct application of the DPO model for answer generation, which yields an accuracy of 68.29\%. This difference suggests the stronger generalization of DPO model in answer generation, despite it is trained on a reward classification loss. The contradictory results to \cite{hosseini2024v} may be due to the difference of tasks, i.e. Math vs. Video QA.
Refer to \cref{apd:dpo} for more results.


\subsection{Analysis on Video Captioning Ability from Pre-training}
\vspace{-0.4cm}
\begin{figure}[ht]
\centering
\begin{tabular}{@{}cc@{}}
\includegraphics[width=0.49\linewidth]{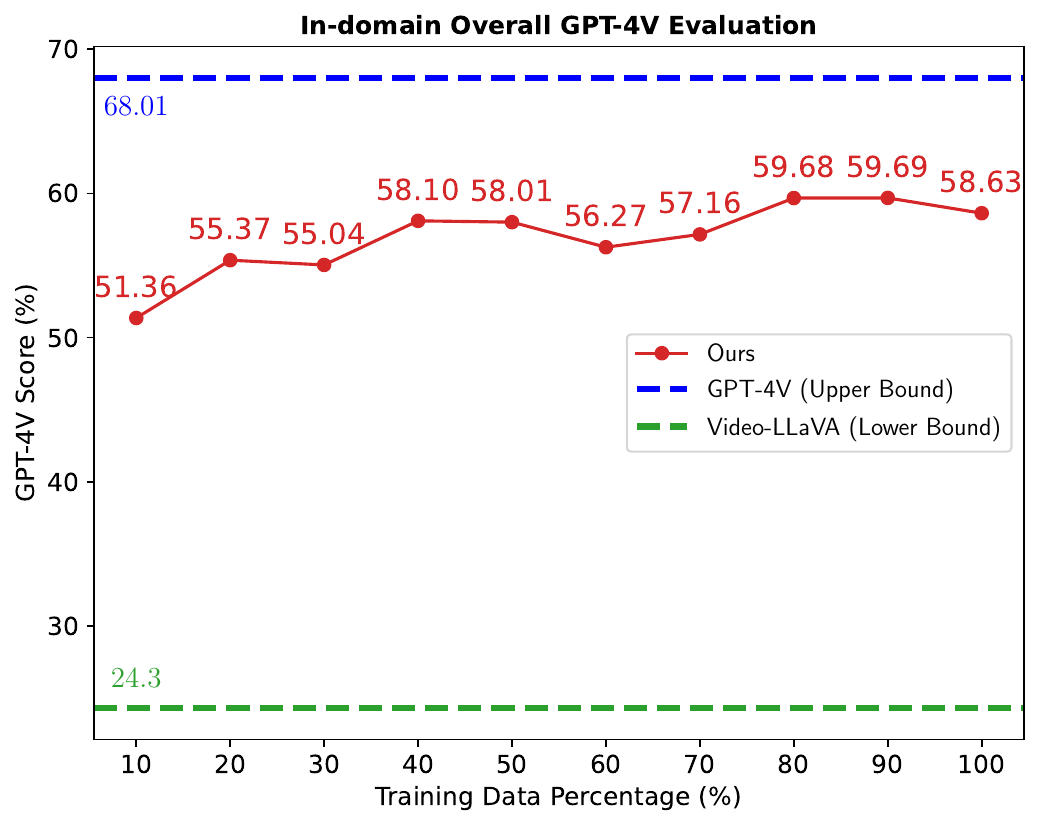} & 
\includegraphics[width=0.49\linewidth]{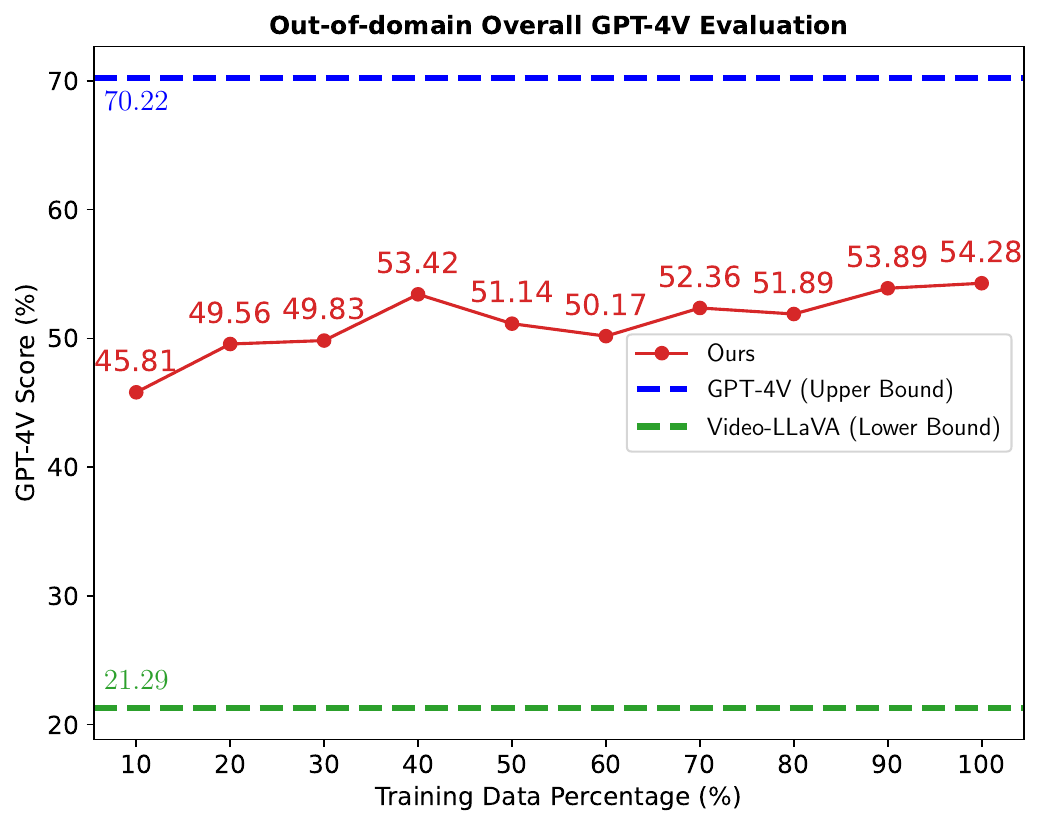} \\
\end{tabular}
\caption{The video caption ability w.r.t number of training data evaluated on both in-domain and out-of-domain test videos using \gptv. 
}
\label{fig:overall_caption_trend}
\end{figure}

In \Cref{fig:overall_caption_trend}, we present the video captioning ability of models across various datasets, with a total of 900k distilled data instances. GPT-4V is employed for self-evaluation (\cref{fig:gpt4v_evaluation}), serving as the upper-bound performance, while the \videollava serves for comparative analysis, establishing a baseline. Notably, \videollava is trained on 54k video QA data instances. However, our first checkpoint, utilizing only 10\% of the data, is trained on 90k high-quality caption data instances, likely accounting for the observed performance disparity in the video captioning task. Our results demonstrate that incorporating more distilled data contributes to improved model performance across both in-domain and out-of-domain datasets. Despite these improvements, a performance discrepancy with the GPT-4V model remains. Further, we evaluate the generalization potential in specific data subsets, as shown in \cref{fig:indomain_caption_trend} in the Appendix. These subsets reveal varying degrees of generalization challenges for different types of dataset. For example, the \webvid subset, which concentrates on relatively static scenes, necessitates less data for effective training compared to the \vidal subset, which is marked by dynamic scene transitions and a diversity of video themes.


\section{Conclusion}
In this study, we propose an cost-effective reward system that utilizes detailed captions as proxies for video content. Our findings demonstrate that the reward scores is well-aligned with the evaluation metrics of \gptv, and the incorporation of this reward mechanism enhances DPO training, resulting in SOTA performance on video QA tasks.

\section{Reproducibility Statement}
The ensure reproducibility of our work, we plan to release the following items:
\begin{enumerate}
    \item Distilled video captions with corresponding frames.
    \item The model weights including the pre-trained, SFT, and DPO models.
    \item Code for training and testing using existing and our proposed benchmark.
\end{enumerate}

\clearpage

\bibliography{colm2024_conference}
\bibliographystyle{colm2024_conference}

\appendix
\section{Effect of \chatgpt Version on Official Benchmark Evaluation}
\label{apd:official_evaluation}

\begin{table*}[!th]
\centering
\resizebox{1.0\linewidth}{!}{
    \begin{tabular}{lccccccccc}
    \toprule
    \multirow{2}{*}{\textbf{Methods}}
    & \multirow{2}{*}{\textbf{LLM Size}}
    & \multicolumn{2}{c}{\textbf{MSVD-QA}}
    & \multicolumn{2}{c}{\textbf{MSRVTT-QA}}
    & \multicolumn{2}{c}{\textbf{TGIF-QA}}  & \multicolumn{2}{c}{\textbf{Summary}} \\
    \cmidrule(lr){3-4}                  
    \cmidrule(lr){5-6}
    \cmidrule(lr){7-8}
    &
    & Acc. & Score 
    & Acc. & Score 
    & Acc. & Score 
    & Avg Acc. & Rank \\
    
    \hline 
    \hline
    & & \multicolumn{6}{c}{gpt-3.5-turbo-0301 evaluation}\\
    \midrule
    Video-ChatGPT~\citep{maaz2023video}
            & 7B & 78.62 & 4.00	& 71.67 & 3.63 & 56.31 & 3.45  & 68.87 & 6 \\
    LLaMA-VID~\citep{li2023llama}
            & 7B & 82.57 & 4.12 & 71.94 & 3.65 & 59.00 & 3.63 & 71.17 & 4 \\
    LLaMA-VID~\citep{li2023llama}
            & 13B & 83.72 & 4.16 & 73.63 & 3.68 & 59.72 & 3.66 & 72.36 & 3 \\
    Chat-UniVi~\citep{Chat-UniVi} & 7B & 80.52 & 4.02 & 66.92  & 3.41 & 57.73 & 3.49 & 68.39 & 7 \\
    Video-LLaVA~\citep{lin2023video}
            & 7B & 81.44 & 4.08 & 73.29 & 3.65 & 58.34 & 3.61 & 71.02 & 5 \\
    \modelsft & 7B & 85.65 & 4.10 & 73.85 & 3.62 & 64.98 & 3.65 & 74.83 & 2 \\
    \modelname & 7B & \bf 88.50 & \bf 4.20 & \bf 82.10 & \bf 3.84 & \bf 75.48 & \bf 3.81 & \bf 82.03 & 1\\

    \hline 
    \hline
    & & \multicolumn{6}{c}{gpt-3.5-turbo-0613 evaluation}\\
    \midrule
    Video-ChatGPT~\citep{maaz2023video}
            & 7B & 68.55 & 3.80 &	58.90 & 3.36 & 47.83 & 3.21 & 58.43 & 6 \\
    LLaMA-VID~\citep{li2023llama}
            & 7B & 72.62 & 3.92 & 58.73 & 3.38 & 49.21 & 3.28 & 60.19 & 4\\
    LLaMA-VID~\cite{li2023llama}
            & 13B & 74.29 & 3.96 & 59.82 &3.41 & 50.83 & 3.33 & 61.65 & 3 \\
    Chat-UniVi~\citep{Chat-UniVi} & 7B & 70.01 & 3.79 & 53.08 & 3.14 & 46.09 & 3.12 & 56.39 & 7 \\
    Video-LLaVA~\citep{lin2023video}
            & 7B & 71.75 & 3.88	& 58.97 & 3.39 & 48.39 & 3.24 & 59.70 & 5 \\
    \modelsft & 7B & 75.70 & 3.86 & 58.73& 3.31	& 53.51 & 3.30 & 62.65 & 2\\
    \modelname & 7B & \bf 80.73 & \bf 4.07 & \bf 70.15 &\bf 3.66 & \bf 61.38 & \bf 3.46 & \bf 70.75 & 1\\

    \hline 
    \hline
    & & \multicolumn{6}{c}{gpt-3.5-turbo-1106 evaluation}\\
    \midrule
    Video-ChatGPT~\citep{maaz2023video}
            & 7B & 73.02 & 4.01	& 62.09 & 3.61 & 47.76 & 3.36 & 60.96 & 6\\
    LLaMA-VID~\citep{li2023llama}
            & 7B & 75.49 & 4.08	& 62.09 & 3.61 & 51.72 & 3.47 & 63.10 & 4 \\
    LLaMA-VID~\citep{li2023llama}
            & 13B & 76.97 & 4.10 & 63.16 &3.61 & 52.53 & 3.50 & 64.22 & 3 \\
    Chat-UniVi~\citep{Chat-UniVi} & 7B & 72.22 & 3.92 & 55.02 & 3.35 & 	48.16 & 3.31 & 58.47 & 7 \\
    Video-LLaVA~\citep{lin2023video}
            & 7B & 74.76 & 4.04 & 62.70 & 3.60 & 51.21 & 3.45 & 62.89 & 5 \\
    \modelsft & 7B & 81.09& 4.08 & 64.13 & 3.57 & 58.05 & 3.53 & 67.76 & 2\\
    \modelname & 7B & \bf 86.05 & \bf 4.23 & \bf 76.75 & \bf 3.85 & \bf 70.02 & \bf 3.71 & \bf 77.61 & 1\\
    
    \bottomrule
    \hline
    \end{tabular}
}
\vspace{-0.5em}
\caption{\textbf{Performance Evaluation Across ChatGPT Versions on Zero-Shot Video Question Answering Benchmarks.} This table compares the performance of state-of-the-art video LMMs evaluated under different \chatgpt versions. The absolute performance metrics scored by \chatgpt vary by versions. However, the comparative ranking of models under the same \chatgpt version is relatively stable.
}
\vspace{-0.5em}
\label{tab:chatgpt_version_official}
\end{table*}

In Table~\ref{tab:chatgpt_version_official}, we show impact of using different ChatGPT versions on metric scores within zero-shot video question answering benchmarks. Our analysis reveals significant variations in the absolute scores across ChatGPT versions, but based on the average accuracy metric, the relative ranking of models under the same ChatGPT version shows consistency.

This comparison underscores a critical issue: many prior studies neglect to specify the ChatGPT version used, potentially leading to inaccurate conclusions during evaluation. We advocate for the explicit designation of the ChatGPT version in future evaluations. Analysis from Table~\ref{tab:chatgpt_version_official} indicates that the version gpt-3.5-turbo-0613 aligns most closely with the performance of the Video-LLaVA~\citep{lin2023videollava} model, serving as the benchmark for model performance comparison in our study.

\section{Evaluation of Captioning Ability from pre-training}
\begin{figure}[ht]
\centering
\begin{tabular}{@{}cc@{}}
\includegraphics[width=0.5\linewidth]{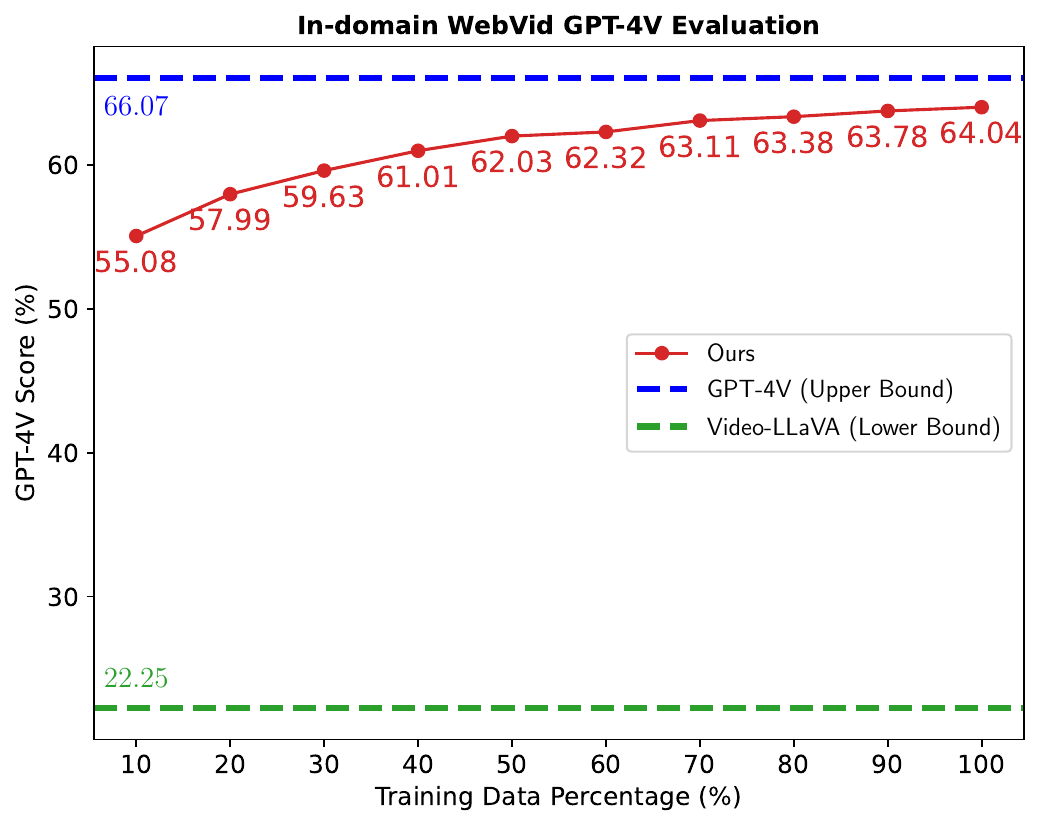} & 
\includegraphics[width=0.5\linewidth]{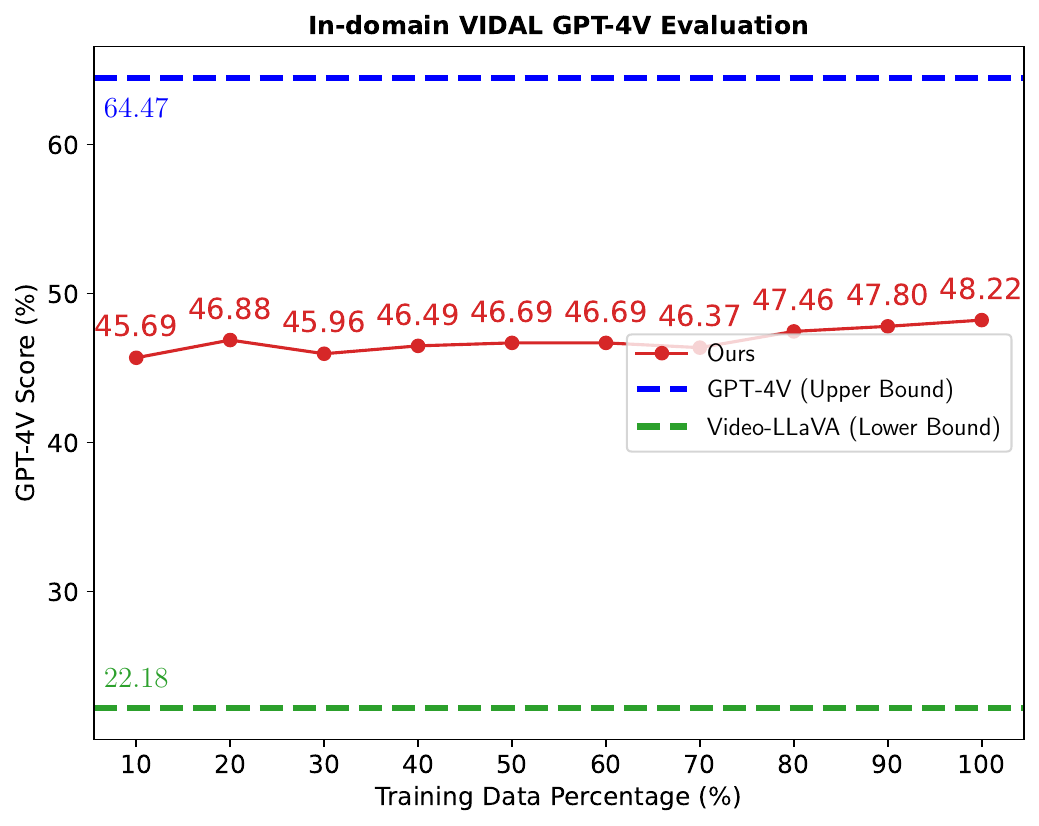} \\
\end{tabular}
\caption{Training subsets exhibit varying levels of generalization difficulty. The WebVid subset (left) requires less data compared to the VIDAL subset (right)}
\label{fig:indomain_caption_trend}
\end{figure}



\section{Distilled Caption Demonstration}
\label{apd:caption}
\begin{figure*}[!ht]
    \centering
    \includegraphics[trim={0cm 5pt 0 5pt},clip, 
    width=1.0\textwidth]{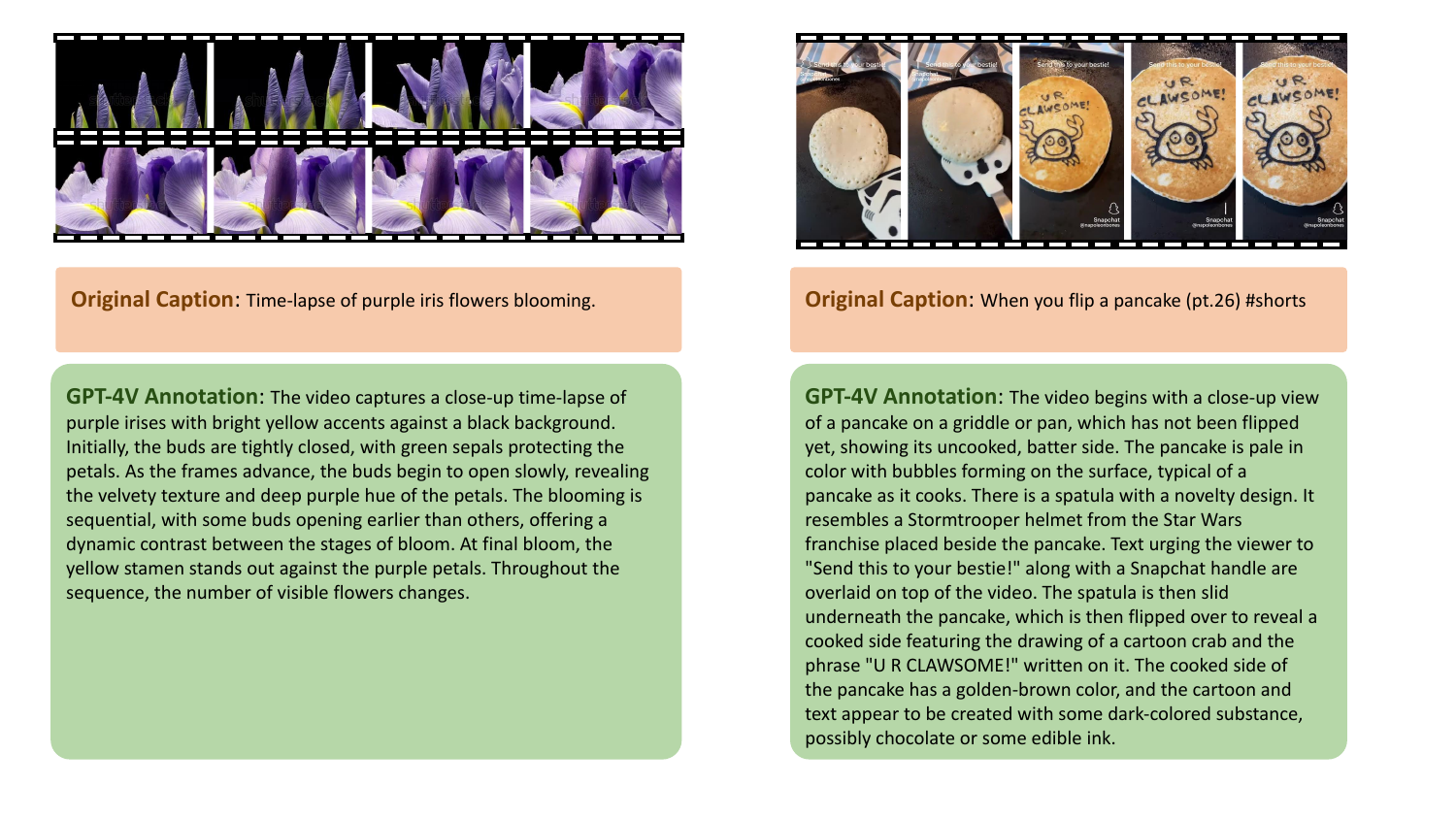}
    \vspace{-2mm}
\caption{Dataset examples annotated by GPT-4V}
\label{fig:dataset-example}
\vspace{-4mm}
\end{figure*}

\section{Video QA Dataset Demonstration}
\label{apd:video_qa}
\begin{figure*}[!ht]
    \centering
    \includegraphics[width=\linewidth]{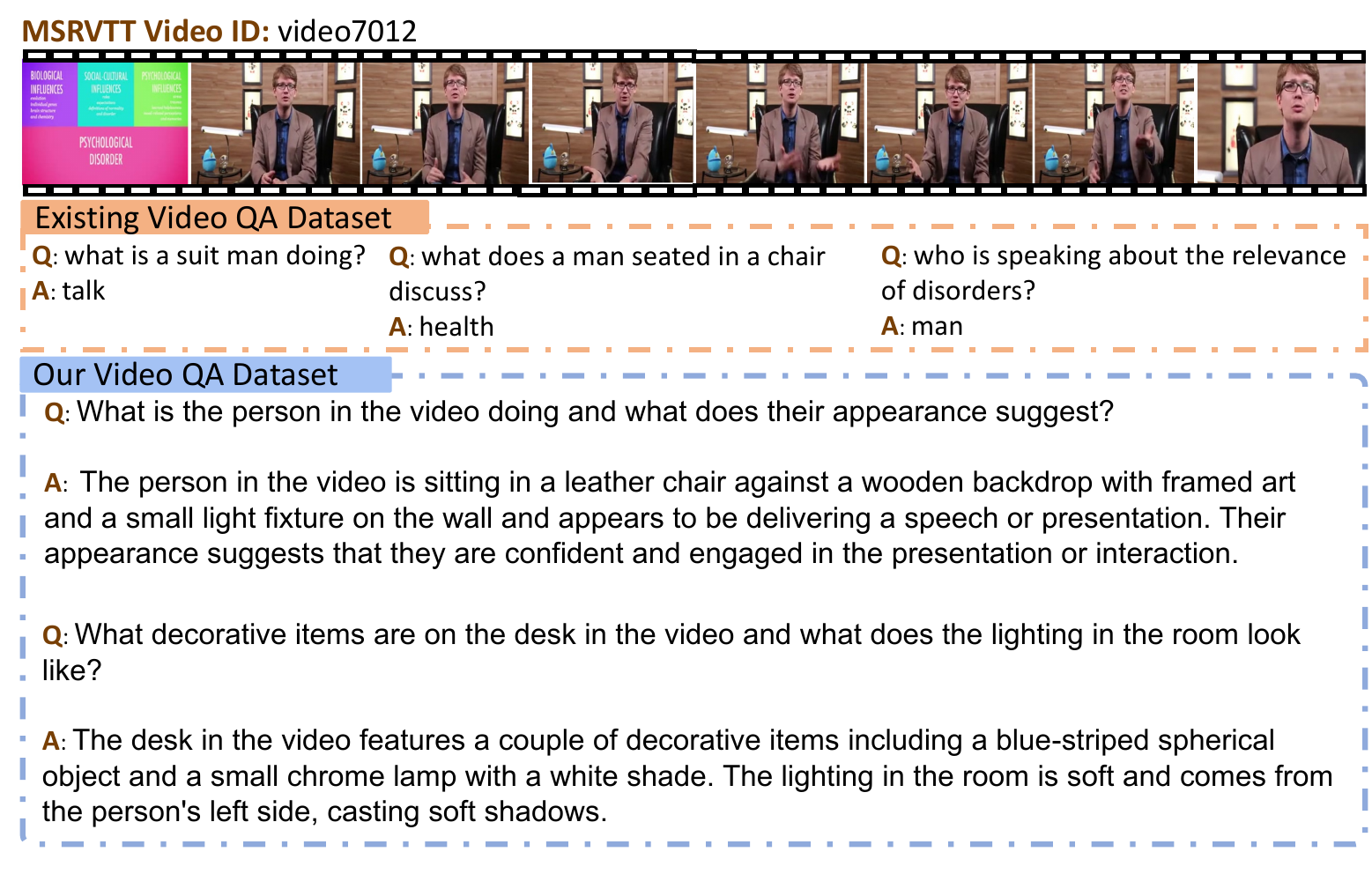}
\caption{Comparing testing QA in existing benchmark with that in our proposed new benchmark.}
\label{fig:dataset_compare}
\end{figure*}

\begin{figure*}[!ht]
    \centering
    \includegraphics[width=\linewidth]{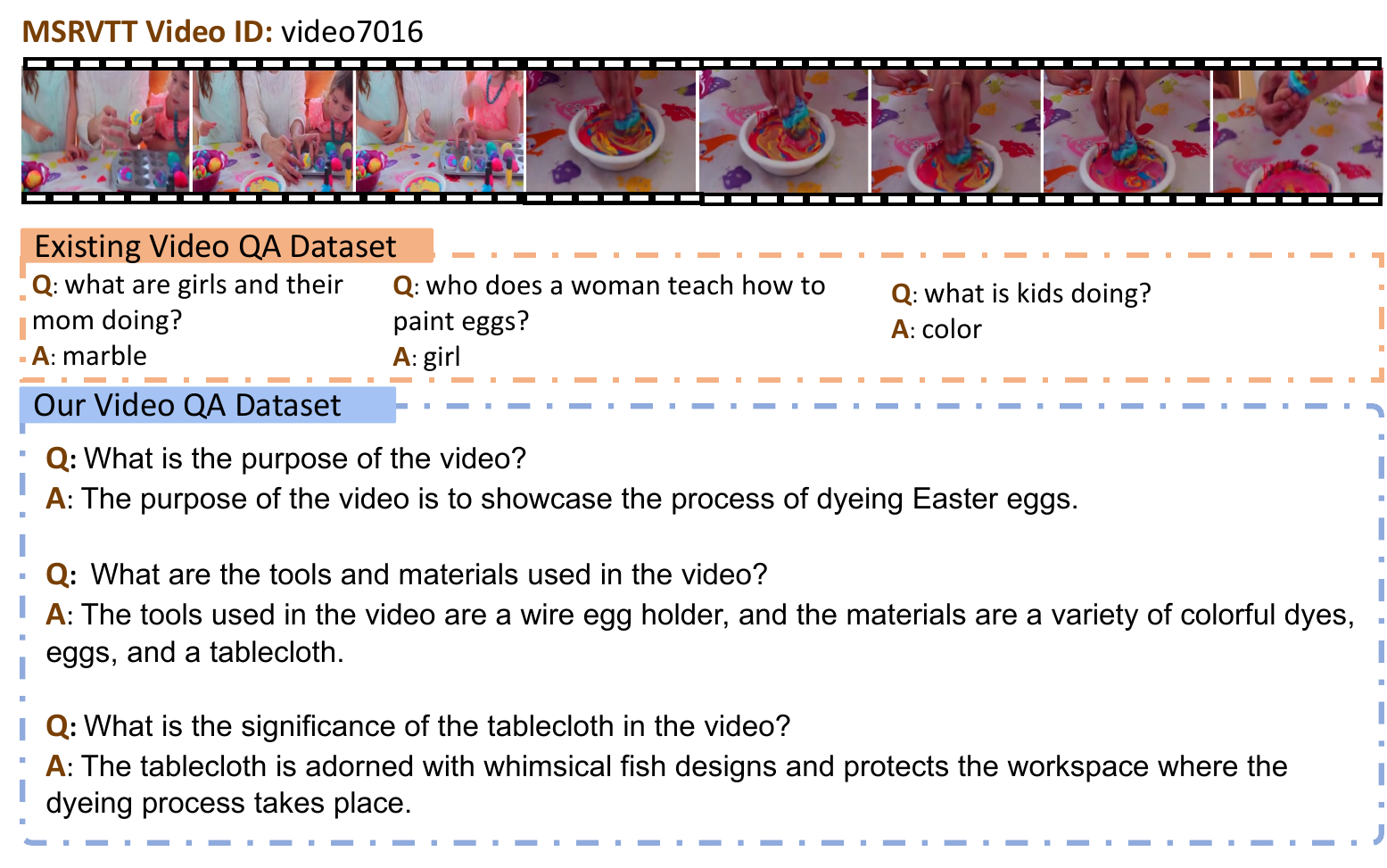}
\caption{Comparing testing QA in existing benchmark with that in our proposed new benchmark, example 2.}
\label{fig:dataset_compare2}
\end{figure*}

\section{Additional DPO Results}
\label{apd:dpo}

\begin{figure}[ht]
\centering
\includegraphics[width=0.49\linewidth]{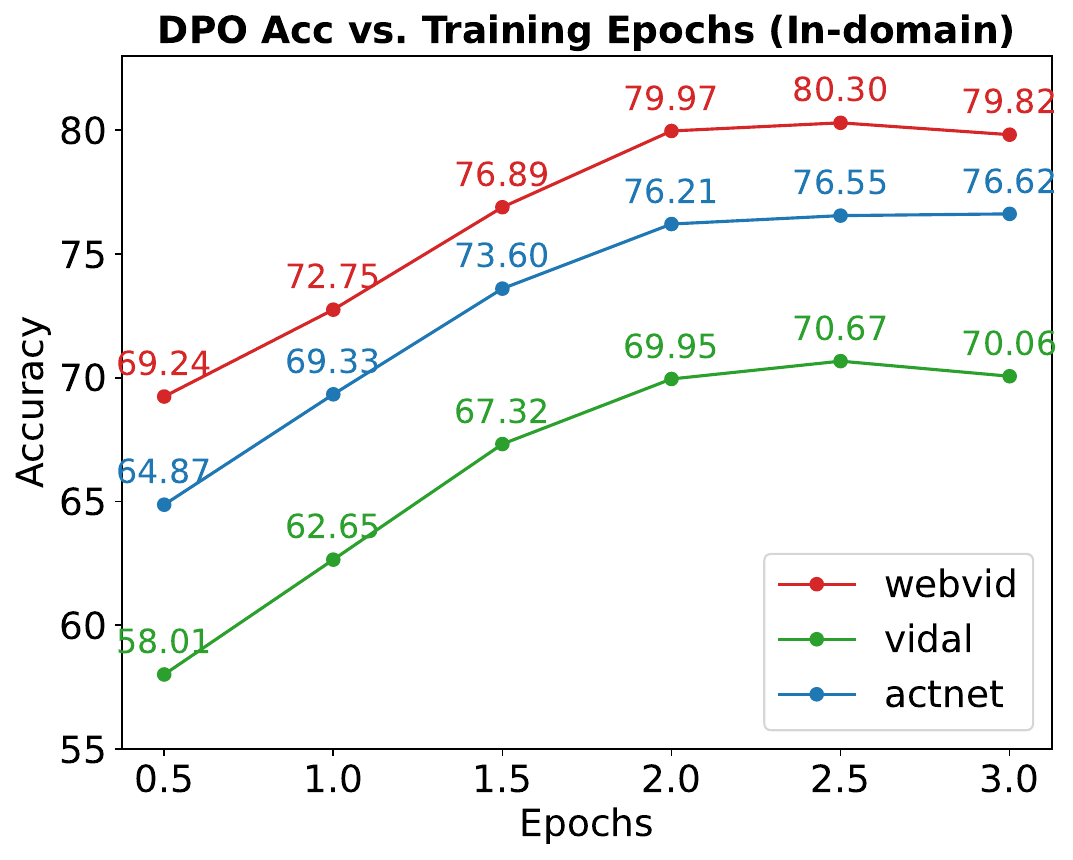} 
\includegraphics[width=0.50\linewidth]{fig/outdomain_dpo.pdf} 
\caption{Test Set Accuracy of the DPO Model vs. Training Epochs. The figure illustrates a consistent trend in both in-domain and out-of-domain video QA, with peak performance occurring at approximately epoch 2.5, equivalent to 350 training steps.}
\label{fig:apd_dpo_epoch}
\end{figure}

\begin{figure}[ht]
\centering
\includegraphics[width=0.50\linewidth]{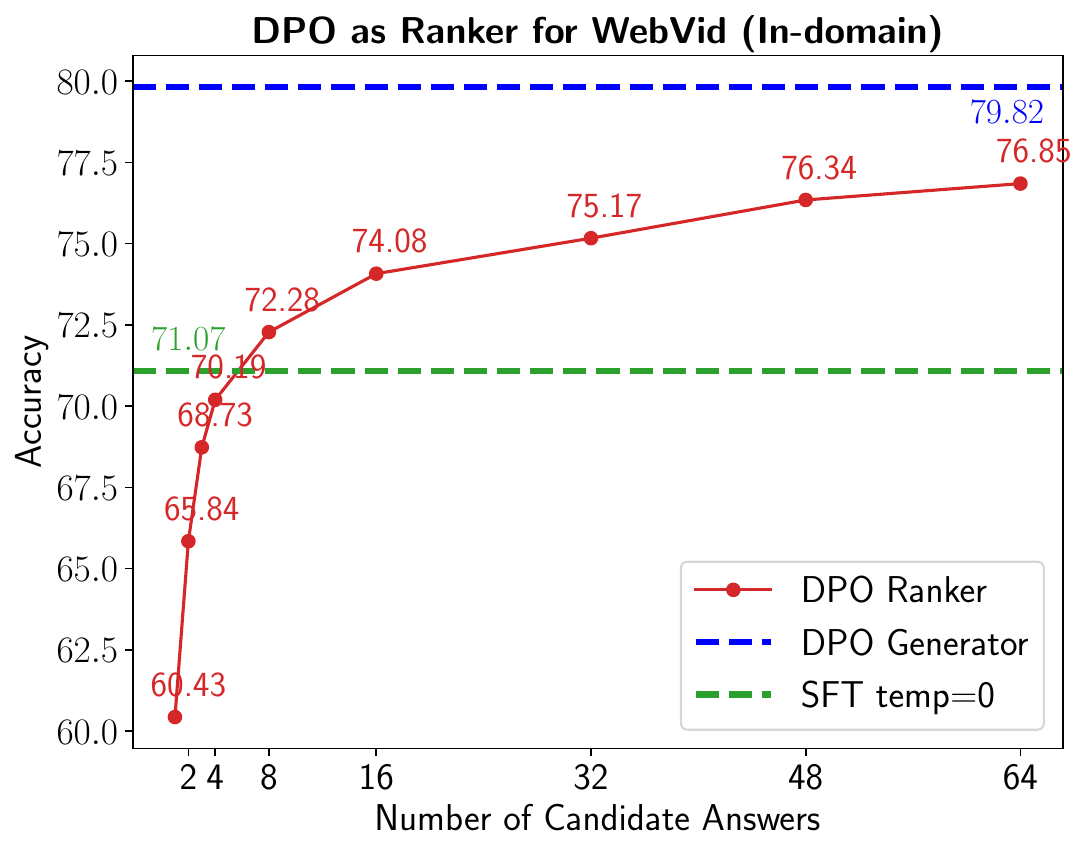} 
\includegraphics[width=0.49\linewidth]{fig/outdomain_dpo_ranker.pdf} 
\caption{Comparison of DPO Model Performance: Ranker vs. Generator. The DPO model serves as a ranker, assigning reward scores to candidate answers generated by the SFT model with a temperature setting of 1.0. Employing the DPO model directly for answer generation results in superior performance compared to its use as a ranker.}
\label{fig:apd_dpo_ranker}
\end{figure}

\section{Prompts for \gptv and \chatgpt Queries}
\label{apd:prompts}

\begin{figure}[ht]
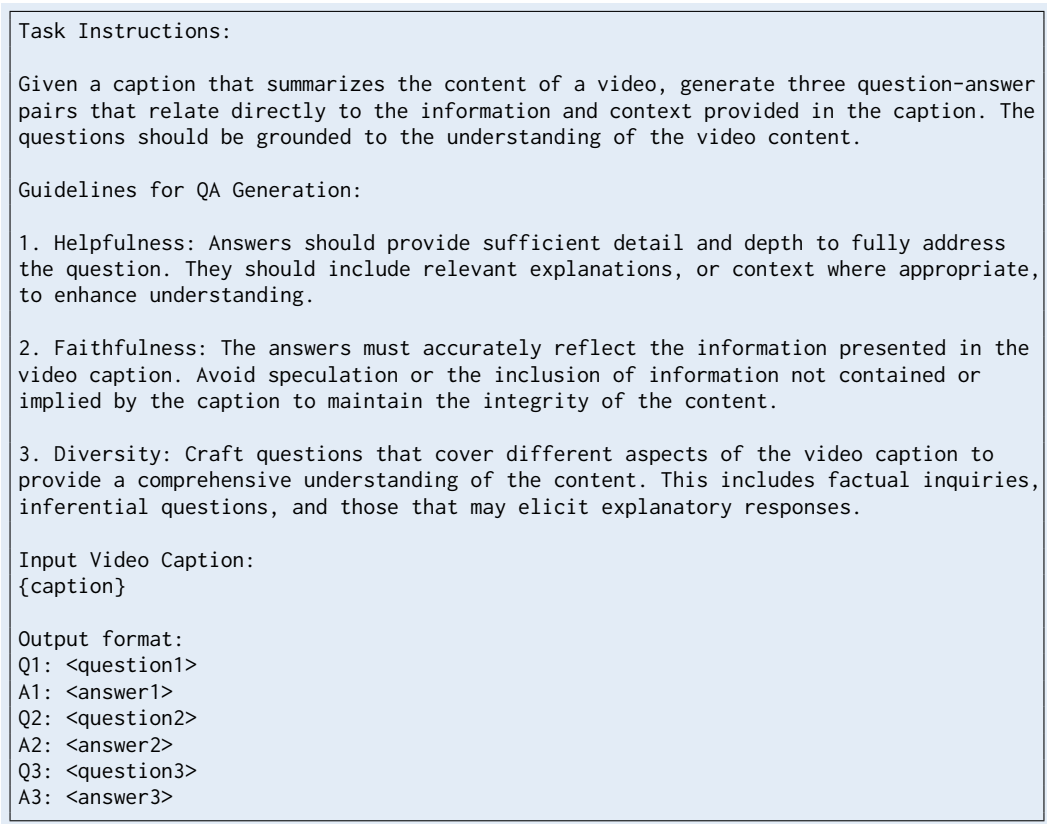

\centering
\small
\centering
\begin{minted}[fontsize=\footnotesize, frame=single,linenos=false,breaklines,breaksymbol=,escapeinside=||,bgcolor=Box1Color]{text}
Task Instructions:

Given a caption that summarizes the content of a video, generate three question-answer pairs that relate directly to the information and context provided in the caption. The questions should be grounded to the understanding of the video content.

Guidelines for QA Generation:

1. Helpfulness: Answers should provide sufficient detail and depth to fully address the question. They should include relevant explanations, or context where appropriate, to enhance understanding.

2. Faithfulness: The answers must accurately reflect the information presented in the video caption. Avoid speculation or the inclusion of information not contained or implied by the caption to maintain the integrity of the content.

3. Diversity: Craft questions that cover different aspects of the video caption to provide a comprehensive understanding of the content. This includes factual inquiries, inferential questions, and those that may elicit explanatory responses.

Input Video Caption:
{caption}

Output format:
Q1: <question1>
A1: <answer1>
Q2: <question2>
A2: <answer2>
Q3: <question3>
A3: <answer3>
\end{minted}
\caption{ChatGPT for instruction generation.}
\label{fig:chatgpt_instruction_generation}
\end{figure}

\begin{figure}[ht]
\centering
\small
\centering
\begin{minted}[fontsize=\footnotesize, frame=single,linenos=false,breaklines,breaksymbol=,escapeinside=||,bgcolor=Box2Color]{text}
Your role is to serve as an impartial and objective evaluator of a video caption provided by a Large Multimodal Model (LMM). Based on the input frames of a video, assess primarily on two criteria: the coverage of video elements in the caption and the absence of hallucinations in the response. In this context, 'hallucination' refers to the model generating content not present or implied in the video, such as incorrect details about objects, actions, counts, or other aspects not evidenced in the video frames.

To evaluate the LMM's response:

Start with a brief explanation of your evaluation process.
Then, assign a rating from the following scale:

Rating 6: Very informative with good coverage, no hallucination
Rating 5: Very informative, no hallucination
Rating 4: Somewhat informative with some missing details, no hallucination
Rating 3: Not informative, no hallucination
Rating 2: Very informative, with hallucination
Rating 1: Somewhat informative, with hallucination
Rating 0: Not informative, with hallucination

LMM Response to Evaluate
{LLM_response}

Output format:
Judgment: <your judgment>
Score: <integer value rating>
\end{minted}
\caption{GPT-4V evaluation prompt for video captioning.}
\label{fig:gpt4v_evaluation}
\end{figure}

\begin{figure}[ht]
\centering
\small
\centering
\begin{minted}[fontsize=\footnotesize, frame=single,linenos=false,breaklines,breaksymbol=,escapeinside=||,bgcolor=Box2Color]{text}
Given the following inputs:

1. **Ground Truth Video Caption**: {caption}
2. **Question Related to the Caption**: {question}
3. **Ground Truth Answer**: {answer}
4. **Model Predicted Answer**: {prediction}

Your task is to evaluate the model's predicted answer against the ground truth answer, based on the context provided by the video caption and the question. Consider the following criteria for evaluation:

- **Relevance**: Does the predicted answer directly address the question posed, considering the information provided in the video caption?
- **Accuracy**: Compare the predicted answer to the ground truth answer. Does the prediction accurately reflect the information given in the ground truth answer without introducing factual inaccuracies?
- **Clarity**: Assess the clarity of the predicted answer. Look for issues such as repetition, unclear descriptions, or any grammatical errors that could hinder understanding.
- **Completeness**: Determine if the predicted answer fully covers the scope of the ground truth answer. Does it leave out critical information or does it include all necessary details?

**Output Format**:
Explanation: <brief judgement of prediction>
Score: <a integer score of quality from 1-5>
\end{minted}
\caption{\chatgpt-Evaluation Prompt for Video Question Answering. This prompt takes in a detailed caption, question, ground truth answer, and model prediction, subsequently generating an assessment of the prediction's quality alongside a corresponding score based on predefined criteria. A score value $\ge 3$ will be considered correct for accuracy calculation. }
\label{fig:chatgpt_verifier}
\end{figure}

\begin{figure}[ht]
\centering
\small
\centering
\begin{minted}[fontsize=\footnotesize, frame=single,linenos=false,breaklines,breaksymbol=,escapeinside=||,bgcolor=Box2Color]{text}
Your task is to act as an impartial and objective assessor of answers generated by a Large Multimodal Model (LMM) for video-based questions. Utilizing video frames, a posed question, and the model's provided answer, your evaluation should focus on the following aspects:

- **Relevance**: Does the predicted answer directly address the question posed, considering the information provided in the video caption?
- **Accuracy**: Compare the predicted answer to the ground truth answer. Does the prediction accurately reflect the information given in the ground truth answer without introducing factual inaccuracies?
- **Clarity**: Assess the clarity of the predicted answer. Look for issues such as repetition, unclear descriptions, or any grammatical errors that could hinder understanding.
- **Completeness**: Determine if the predicted answer fully covers the scope of the ground truth answer. Does it leave out critical information or does it include all necessary details?

**Input**:
Question: {question}
Model Predicted Answer: {prediction}

**Output Format**:
Explanation: <brief judgement of prediction>
Score: <an integer score of quality from 1-5>
\end{minted}
\caption{\gptv Evaluation Prompt for Video Question Answering. Together with video frames input in \gptv API, this prompt takes in a question, and model prediction, subsequently generating an assessment of the prediction's quality alongside a corresponding score based on predefined criteria. A score value $\ge 3$ will be considered correct for accuracy calculation. This is used to assess the quality of \chatgpt evaluation in \cref{fig:chatgpt_verifier}. }
\label{fig:gptv_verifier}
\end{figure}

\end{document}